%% file: main.tex
\documentclass[conference]{IEEEtran}
\IEEEoverridecommandlockouts
\makeatletter
\newcommand{\linebreakand}{%
  \end{@IEEEauthorhalign}
  \hfill\mbox{}\par
  \mbox{}\hfill\begin{@IEEEauthorhalign}
}
\makeatother

\usepackage{cite}
\usepackage{amsmath,amssymb,amsfonts}

\usepackage{graphicx}
\usepackage{textcomp}
\usepackage{xcolor}
\usepackage{algpseudocode}
\usepackage{algorithm}
\usepackage{amssymb}
\usepackage{mathrsfs}
\usepackage{subcaption}
\def\BibTeX{{\rm B\kern-.05em{\sc i\kern-.025em b}\kern-.08em
    T\kern-.1667em\lower.7ex\hbox{E}\kern-.125emX}}
\begin{document}

\title{Fool the Hydra: \\ Adversarial Attacks against Multi-view Object Detection Systems}

\author{\IEEEauthorblockN{Bilel Tarchoun}
\IEEEauthorblockA{\textit{LATIS Laboratory} \\
\textit{Ecole Nationale d’Ingénieurs de Sousse}\\
Sousse, Tunisia \\
bilel.tarchoun@eniso.u-sousse.tn}
\and
\IEEEauthorblockN{Quazi Mishkatul Alam}
\IEEEauthorblockA{\textit{University of California Riverside} \\
California, USA\\
quazimishkatul.alam@email.ucr.edu}
\and
\IEEEauthorblockN{ Nael Abu-Ghazaleh}
\IEEEauthorblockA{\textit{University of California Riverside} \\
California, USA \\
nael.abughazaleh@ucr.edu}
\linebreakand
\IEEEauthorblockN{ Ihsen Alouani}
\IEEEauthorblockA{\textit{CSIT,  Queen's University} \\
Belfast, UK \\
i.alouani@qub.ac.uk}
}

\maketitle

\begin{abstract}
Adversarial patches exemplify the tangible manifestation of the threat posed by adversarial attacks on Machine Learning (ML) models in real-world scenarios. Robustness against these attacks is of the utmost importance when designing computer vision applications, especially for safety-critical domains such as CCTV systems. In most practical situations, monitoring open spaces requires multi-view systems to overcome acquisition challenges such as occlusion handling. Multiview object systems are able to combine data from multiple views, and reach reliable detection results even in difficult environments. Despite its importance in real-world vision applications, the vulnerability of multiview systems to adversarial patches is not sufficiently investigated. In this paper, we raise the following question: Does the increased performance and information sharing across views offer as a by-product robustness to adversarial patches? We first conduct a preliminary analysis showing promising robustness against off-the-shelf adversarial patches, even in an extreme setting where we consider patches applied to all views by all persons in Wildtrack benchmark. However, we challenged this observation by proposing two new attacks: \textbf{(i)} In the first attack, targeting a multiview CNN, we maximize the global loss by proposing gradient projection to the different views and aggregating the obtained local gradients. \textbf{(ii)} In the second attack, we focus on a Transformer-based multiview framework. In addition to the focal loss, we also maximize the transformer-specific loss by dissipating its attention blocks. 
Our results show a large degradation in the detection performance of victim multiview systems with our first patch attack reaching an attack success rate of $73\%$ , while our second proposed attack reduced the performance of its target detector by $62\%$
\end{abstract}


\input{sec/1_intro}

\input{sec/2_preliminaries}
\input{sec/3_mvdet_attack}

\input{sec/4_mvdetr_attack}
\input{sec/5_discussion}
\input{sec/6_related}

\input{sec/7_conclusion}

\bibliographystyle{IEEEtran}
\bibliography{main.bib}
\vspace{12pt}

\end{document}

%% file: sec/1_intro.tex
\section{Introduction}
\label{sec:intro}

The threat posed by adversarial attacks against ML-powered applications has been extensively studied by the scientific community \cite{surveyaudio,surveycv,surveynlp}.  These threats are particularly prominent against computer vision systems, as a nearly imperceptible adversarial noise can fool object detectors and classifiers \cite{szegedy_noise,FGSM,BIM,cw}. In different settings, adversarial patches can be crafted such that they can fool ML-based computer vision systems under real-life constraints, by concentrating the adversarial noise within a localised area implementable in the physical world \cite{advpatch,yolopatch,naturalistic}. To mitigate physical patch attacks, many defenses have been proposed; two primary categories of defenses either detect and remove the offending noise \cite{lgs,jedi,jujutsu}, or immunize the detector against certain categories of attacks while providing certifiable bounds \cite{patchcleanser,vitcert,smoothedvit}. These defenses however are not perfect; they can result in utility loss, or ultimately new advanced attacks are able to bypass them.

Multiview video analytics is a prominent computer vision technology that is crucial to several application domains such as object detection and action recognition in complex environments such as outdoor or crowded spaces \cite{mvdet,cnncrf,vfa}. Multiview systems are incentivized by the limitations of single view models. For example, single-view performance is limited in occlusion handling: fully occluded objects are impossible to detect, while partially occluded objects are challenging to properly identify, detect or recognize. Multiview systems overcome this limitation by  sharing the information contained within multiple views to improve their performance. Despite the importance of multiview systems, their robustness against adversarial attacks is not well explored.  

In this paper, we investigate the robustness of multiview systems against adversarial patches under real-world settings. We first ask the question: \textbf{Does a multiview setting offer inherent robustness against adversarial patches?} We
carry out a preliminary study of the efficiency of existing off-the-shelf adversarial patches in fooling state-of-the-art multiview detectors. Interestingly, we discover that multiview systems are robust against these attacks, even when tested under extreme assumptions with many adversarial patches attacking the individual views. To challenge this observation and further explore the robustness of these systems, we propose two novel adversarial patch attacks that are specifically crafted for multiview object detectors. These patches leverage data from all of the views of the victim system in order to generate an effective adversarial patch. 

The first patch attack projects the gradient resulting from the detectors loss back to each view separately. These gradients are used to extract local view-specific gradients that are relevant to each instance of patch placement. These local gradients are aggregated to form the update step of the patch and thus generate a patch that targets all of the views of the victim system. We show that by aggregating gradient projection in different views, the adversarial noise succeeds in fooling the multiview system.  However, we found that this attack does not transfer to vision transformer based systems.  Therefore, we develop a second patch attack which targets multiview detectors that include transformers in their frameworks. By refactoring the patch generation process to maximize the focal loss along with the regular detection loss, this patch successfully fools a multiview detector that applies a shadow transformer to the features of the input images and reached an attack success rate of $62\%$.  Our results demonstrate that despite the apparent robustness shown by against existing adversarial attacks, new multi-view specific patches are effective, proving that multiview detectors are indeed vulnerable to adversarial attack.

Our contribution can be summarized as follows:

\begin{itemize}
    \item To our knowledge, we are first to investigate the adversarial robustness of multi-view systems under real-world constraints. We show that these systems have a level of inherent robustness because of their complementary multi-stream architecture. 
    
    \item We propose an adversarial patch attack that aggregates data from multiple views to generate multiview-specific adversarial noise, and we show that this attack does not transfer to transformer-based multiview architectures. 
    
    \item  We propose a second adversarial patch attack that can attack transformer-based detectors by maximizing a loss function that undermines the global system and dissipates the attention mechanism simultaneously. 
\end{itemize}

%% file: sec/2_preliminaries.tex
\section{Does multi-view help defending against adversarial patches?}
\label{sec:prelim}

In this section, we propose a preliminary analysis where we evaluate the performance of existing single view adversarial patches against a state-of-the-art multiview object detector.

\noindent\textbf{Setup.} Our first experiment consists of applying single view adversarial patches to the Wildtrack dataset \cite{wildtrack} in order to attack the MVDET \cite{mvdet} multiview detector. The Wildtrack dataset is a large dataset containing seven different views with high image quality that film a courtyard where a crowd of people gather and move without restriction. The synchronized and calibrated views along with the unrestricted flow of persons accurately represents the challenges encountered in difficult object detection tasks, making the Wildtrack dataset a perfect candidate for simulating a real-world implementation of a CCTV system. We combine this dataset with the MVDET detector as it is a state of the art multiview object detector that achieves high performance using a fully convolutional architecture to perform the inter-view feature aggregation.

\begin{figure}
    \centering
    \includegraphics[width = \columnwidth ]{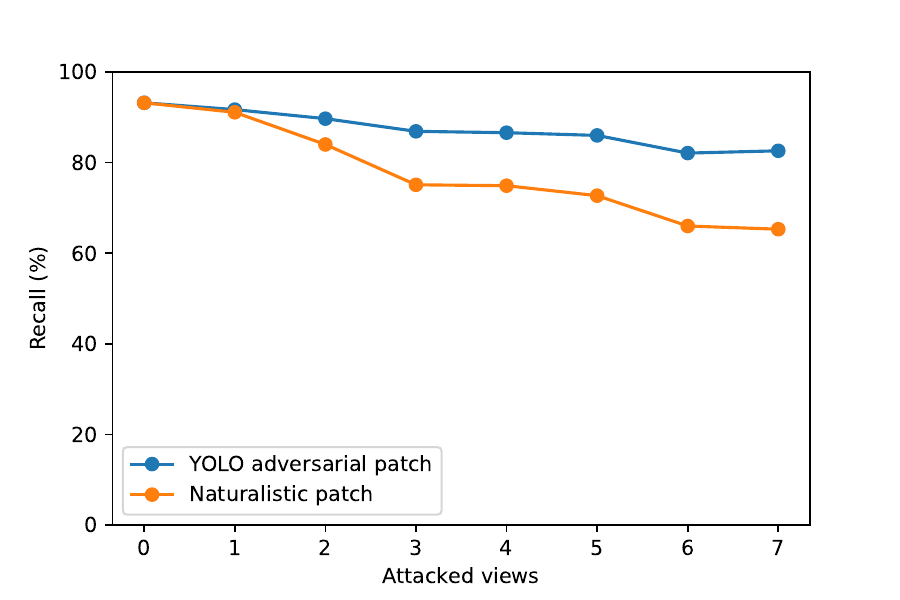}
    \caption{Results of single view patches against MVDet according to the number of attacked views}
    \label{fig:sp1}
\end{figure}

We incrementally  increase the number of views that are systematically attacked until all views of the Wildtrack datasets are attacked. The results for the attacks using YOLO adversarial patch \cite{yolopatch} and Naturalistic patch  \cite{naturalistic} are reported in figure \ref{fig:sp1}

As expected, the success of the adversarial attack increases with the number of attacked views. However, even when all views are attacked, the overall performance of the adversarial patch is low.

\begin{figure}[h]
    \centering
    \includegraphics[width = \columnwidth ]{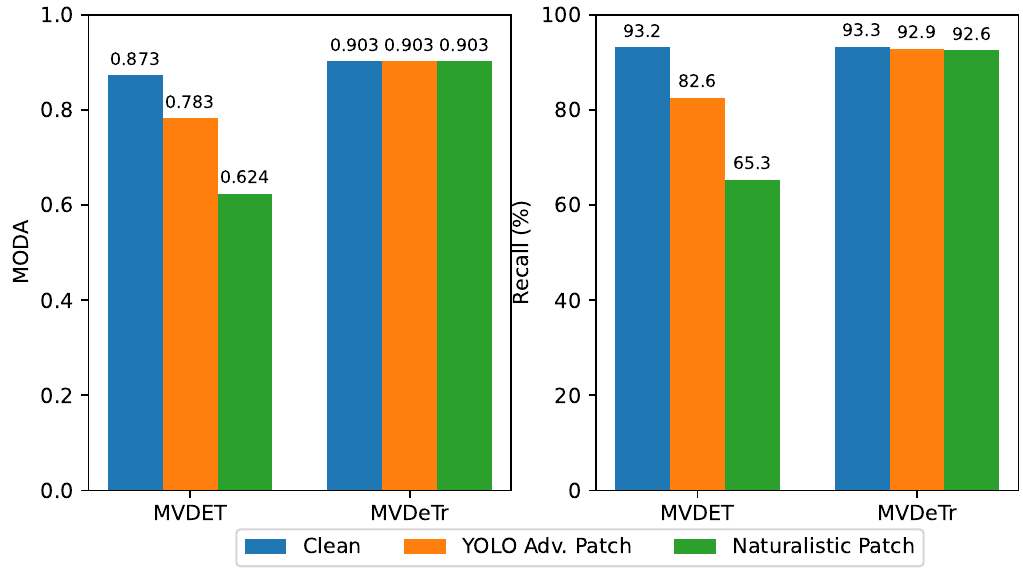}
    \caption{Results of single view patches against other multi view object detectors}
    \label{fig:sp2}
\end{figure}

In our second experiment, we also attack MVDeTr \cite{mvdetr} with the single view patches. MVDeTr improves upon MVDET's performance by including a shadow transformer to account for the projection distortions. The use of non-CNN elements in multi-view object detection is common, therefore it is important to evaluate how adversarial patches perform against such detectors. The results shown in figure \ref{fig:sp2} confirm the trends of the previous experiments: Single view adversarial patches are completely ineffective against MVDeTR.

These results confirm that current single view adversarial patches are ineffective against multi view detectors, This is likely due to the data sharing and information fusion between views that these detectors include in their frameworks. To fool these detectors, a multi view patch is needed: This patch must be able to learn from multiple views of a scene at once in order to be able to fool all targeted cameras simultaneously. 

%% file: sec/3_mvdet_attack.tex
\begin{figure*}[!htp]
    \centering
    \includegraphics[width = \textwidth ]{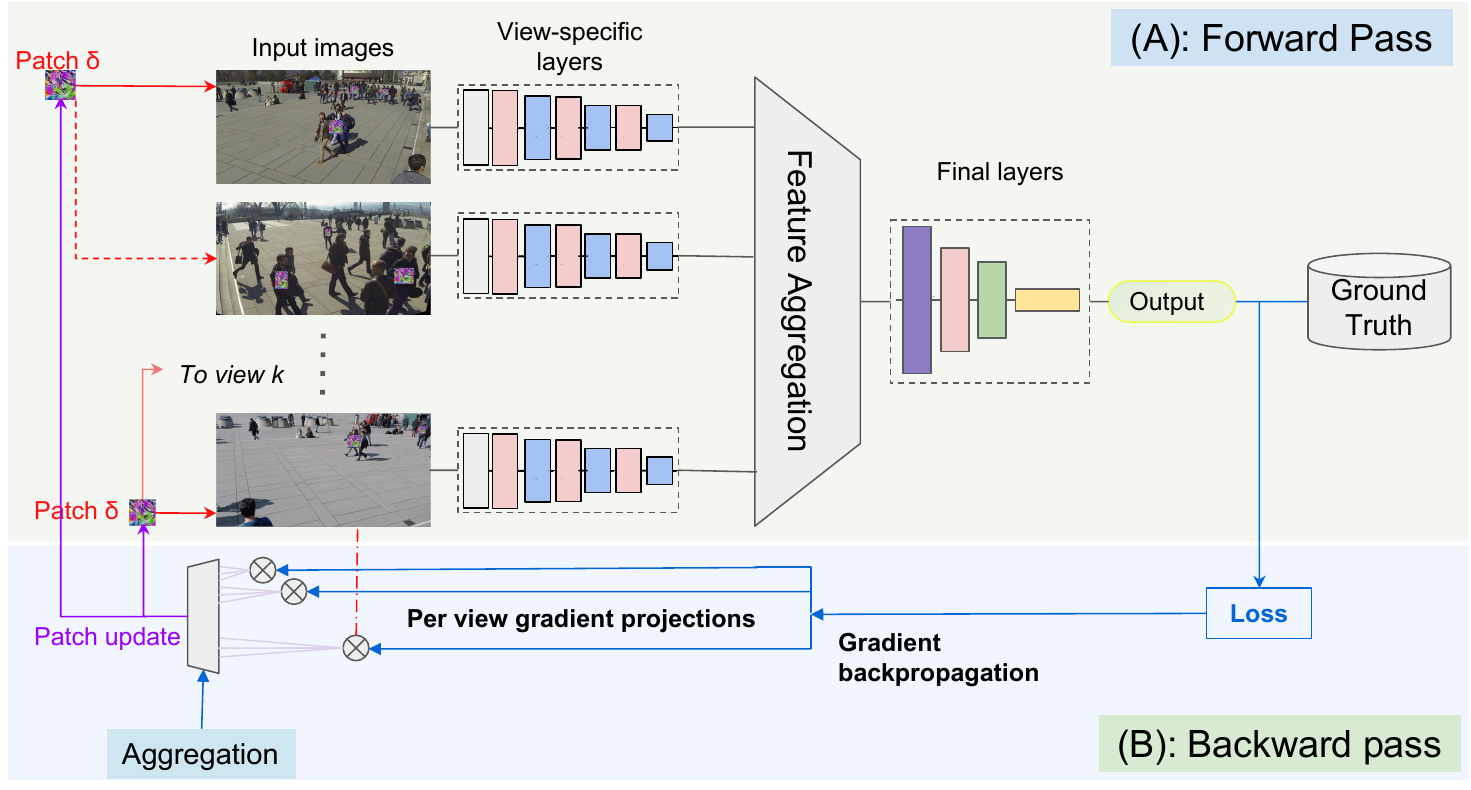}
    \caption{Overview of our proposed multiview patch generation framework}
    \label{fig:mv_schematic}
\end{figure*}

\section{Fool the Hydra: Adaptive attacks against multi-view systems}
\label{sec:approach_1}

In this section, we present our patch attack against multiview object detectors. We define a multi view detector as $f(sv(.))$ where $sv(.)$ is a function that processes each input image in a multiview system with $n$ views $\{x_1,x_2,...,x_n\}$ to extract single view features / information and $f(.)$ is a function that aggregates the information obtained by applying the $sv(.)$ function to all input images in order to obtain the final detection results $O = f(sv(x_1),sv(x_2),...,sv(x_n))$. 

\subsection{Multiview Patch}
Our proposed attack backpropagates the detector's loss and projects the resulting gradient back to each of the views. 
We defined the loss as the combination of two cost functions, following the state-of-the-art in multiview detection \cite{mvdet}: %
\begin{itemize}
    \item A ground plane loss that calculates the difference between the detectors output $\bar{g}$ and the ground truth $g$ using the euclidean distance: 
    \begin{equation}
        \mathscr{L}_{ground} = \left \| \bar{g} - g \right \|_2
    \end{equation}
    \item Single view detection loss that for each view calculates the euclidean distance between the head and foot positions of the detected pedestrians $(\bar{s}_{head}^{(n)},\bar{s}_{foot}^{(n)})$ and those positions in the ground truth $ (s_{head}^{(n)},s_{foot}^{(n)})$
    \begin{equation}
        \mathscr{L}_{single}^{(n)} = \left \|  \bar{s}_{head}^{(n)} - s_{head}^{(n)}\right \|_2 + \left \|  \bar{s}_{foot}^{(n)} - s_{foot}^{(n)}\right \|_2
    \end{equation}
\end{itemize}
The total loss is then calculated as:
\begin{equation}
    \mathscr{L}_{multiview} = L_{ground} + \omega \times \frac{1}{N} \sum_{n=1}^N L_{single}^{(n)}
\label{eq:mult_loss}
\end{equation}
Where $\omega$ is a weight that controls the importance of single view detection in the loss.
The relevant gradient data for each patch application is then aggregated to update the patch according to Algorithm \ref{alg1}.

\begin{algorithm}
    \caption{Multiview patch generation algorithm}
    \label{alg1}
    \begin{algorithmic}[1]
        \State \textbf{Input:} {$patch\_pos\_list$: list of positions of patch targets, $x_1,...,x_N$ input image for view $1,...,N$, $psize$: patch size, $p_{i,v}$: instance $i$ of patch placement on view $v$}
        \State \textbf{Output:} {$\delta$: Adversarial multiview patch}
        \State \textit{/* Start with an initial gray patch */}
        \State { $\delta$ $\xleftarrow{}$ \textit{{patchInit}}(gray,psize)}
        \For {$e \in [1,n_{epochs}]$}
            \For {$it \in [1,n_{iter}]$}
                \State \textit{/* Place the patch on its targets */}
                \State { $x_1,...,x_N$ $\xleftarrow{}$ \textit{{placePatch}}($x_1,...,x_N$ ,$patch\_pos\_list$,$\delta$)}
                \State \textit{/* Run the detector and get the loss */}
                \State {$\mathscr{L}$ $\xleftarrow{}$ \textit{{det($x_1,...,x_N$)}}} 
                \State \textit{/* Project the gradient to each view */}
                \State { $\nabla x_1,...,\nabla x_N$ $\xleftarrow{}$ $\frac{\partial \mathscr{L}}{\partial x_1} ,..., \frac{\partial \mathscr{L}}{\partial x_N}$ }
                \State \textit{/* Obtain the gradient for each patch placement*/}
                \State { $\nabla p_{i,v}$ $\xleftarrow{}$ $\nabla x_v(patch\_pos\_list(i,v))$}
                \State \textit{/* Interpolate the obtained gradients into size $psize$*/}
                \State { $\nabla p_{i,v}$ $\xleftarrow{}$ \textit{{interpolate}}($\nabla p_{i,v}$ , $psize$)}
                \State \textit{/* Aggregate the gradients into the patch update values*/}
                \State { $step$ $\xleftarrow{}$  $\frac{\sum_{v}\sum_{i} \nabla p_{i,v}}{length(patch\_pos\_list)}$}
                \State \textit{/* Update the patch*/}
                \State {$\delta(it+1) = \delta(it) + \alpha * step$}
            \EndFor
        \EndFor
    \end{algorithmic}
\end{algorithm}

Starting from an initial state of a gray patch $\delta$, we place the adversarial patch(es) on their intended targets on each view, we note the position of instance $k$ of patch placement on view $v$ as $p(k,v) = (px_{min}^k,px_{max}^k,py_{min}^k,py_{max}^k)$ and $m_v$ the total number of patches placed on view $v$. 
To account for the differences in view angles between the cameras and their effects on the patch, we use geometric transformations to transfer the patch across views: We split the views in the scene in half, ensuring that the cameras are divided into two sets that are facing each other. One camera in each set is designated as a source camera, where the patches are placed directly . The other cameras are designated as destination cameras, in these views the patches are transferred from the source camera using  geometric transformations. The parameters from these transformations can be calculated using camera calibration data or provided annotations. 
This set of images with the current iteration of patch applied to them is then used as input for the detector. The loss obtained using the loss function $L$ after running the detector is backpropagated through the detector's framework and projected to each view in order to obtain the per-view gradients. The gradient of view $v$ is calculated as follows:

\begin{equation}
    \nabla x_v = \frac{\partial L(f(sv(x_1),sv(x_2),...,sv(x_n))))}{\partial x_v}
    \label{eq:proj_grad}
\end{equation}

The gradients relevant to each patch application can then be extracted from the per-view gradients:

\begin{equation}
    \nabla p(k,v) = \nabla x_v (px_{min}^k,px_{max}^k,py_{min}^k,py_{max}^k)
\end{equation}

As these gradients are of different sizes, we use interpolation is order to standardise the sizes to be equal to the patch(es) size. We then aggregate the various gradients to form the patch update step values.

\begin{equation}
    step = \frac{ \sum_{v=1}^{n} \sum_{k=1}^{m_v} \nabla p(k,v)}{\sum_{v=1}^{n} m_v}
\end{equation}

The patch is then updated using the calculated values : 

\begin{equation}
    \delta(it+1) = \delta(it) + \alpha * step
\end{equation}

Where $\alpha$ is a gradient amplification value that controls the evolution speed of the patch.

This loop is repeated until the appropriate stop reason is reached such as a high loss or maximum epochs.

\subsection{Experimental results}
\textbf{Experimental Setup:} 

We evaluate our patch to two different multiview object detectors:
\begin{itemize}
    \item \textbf{MVDET \cite{mvdet}:} This detector first extracts single view features from each input image using a ResNet-18 CNN architecture, and projects these features to the ground plane using a perspective geometric transformation whose parameters can be calculated from camera calibration data. To perform multiview aggregation while conserving spatial consistency and correctly disambiguating the features of neighboring persons, the authors use a final block of convolutional layers with a large receptive field to process the combined feature maps and output the final pedestrian occupation map on the ground plane.
    \item \textbf{MVDeTr \cite{mvdetr}:} An improvement over MVDET, MVDeTr implements a shadow transformer to deal with the distortions introduced by the ground plane projection operation:  Similarly to its predecessor, MVDeTr uses ResNet-18 to extract features from each input image and then projects these features to the ground plane using a perspective geometric transformation. MVDeTr then applies a multiview deformable attention-based transformer that simultaneously considers all views to the projected features in order to link feature cues across camera views. These enhanced projected features are then used to generate the pedestrian occumancy map using a final block of convolution layers. 
\end{itemize}

We test these detectors with the Wildtrack dataset \cite{wildtrack}, which contains seven \textbf{synchronised} cameras around a courtyard where people's movements are unrestricted. Wildtrack is the state-of-the-art benchmark that enables experimenting multiview CCTV system under real life settings.

In our experiments, we apply our generated multiview adversarial patch to every person present in all of the seven views of Wildtrack.

\begin{figure}[h]
    \centering
    \includegraphics[width = \columnwidth ]{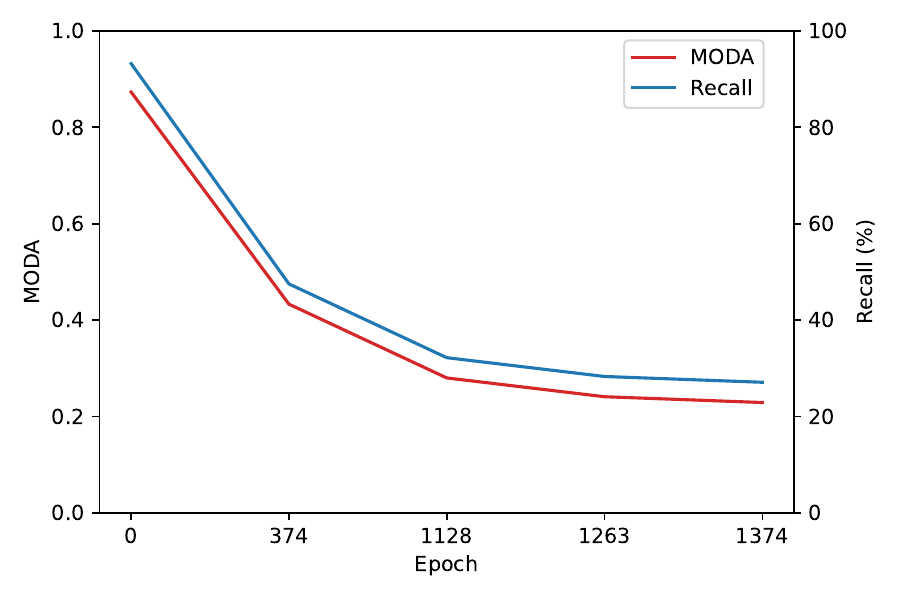}
    \caption{Results of our proposed multiview patch against MVDET}
    \label{fig:res_mvdet}
\end{figure}

We evaluate our patch against MVDET using the Wildtrack dataset, the results across certain checkpoints during patch generation are shown in figure \ref{fig:res_mvdet}. Our patch was able to reduce the original detector's MODA by $-73.77\%$ and the recall by $-70.92\%$ . Figure \ref{fig_ground} shows a sample of MVDET's degraded detection abilities using a heatmap of detections on the ground plane:  Significantly less persons are detected when the patch is applied.

\begin{figure}[h]
    \centering
    \includegraphics[width = \columnwidth ]{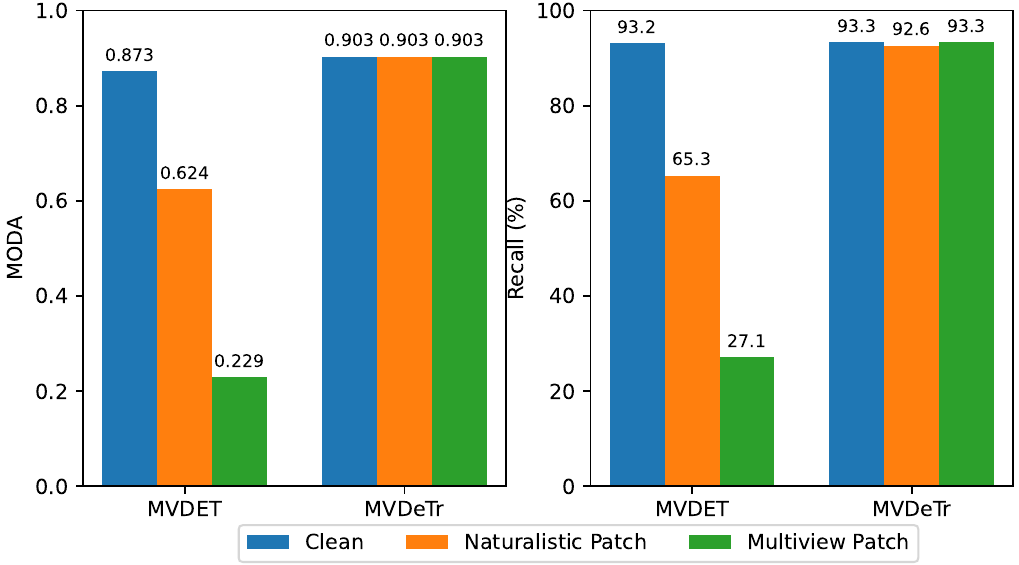}
    \caption{Results of our patch against other multiview detectors }
    \label{fig:res_mvdet_on_other}
\end{figure}

Our patch has successfully fooled MVDET. However, when evaluated against other multiview detectors, its performance lowers drastically, as shown in figure \ref{fig:res_mvdet_on_other}.

We notice that the patch is not transferable to other multiview detectors, likely due to the difference between the frameworks of the detectors and the inclusion of non-CNN elements. Therefore other detectors need attacks specific to them. to this aim we propose a second attack:

\begin{figure*}
     \centering
     \begin{subfigure}[b]{0.24\textwidth}
         \centering
         \includegraphics[width=\textwidth]{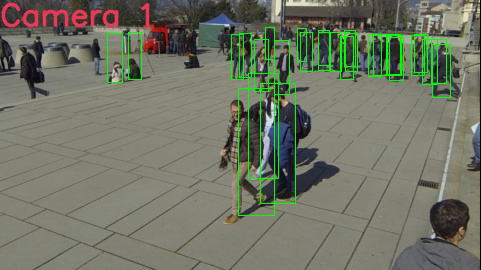}
         \caption{Camera 1 view: No patch}
         \label{cam1_cleam}
     \end{subfigure}
     \hfill
     \begin{subfigure}[b]{0.24\textwidth}
         \centering
         \includegraphics[width=\textwidth]{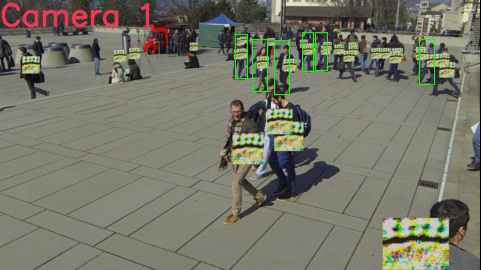}
         \caption{Camera 1 view: Patch Applied}
         \label{cam1_patch}
     \end{subfigure}
     \hfill
     \begin{subfigure}[b]{0.24\textwidth}
         \centering
         \includegraphics[width=\textwidth, height= 2.36cm]{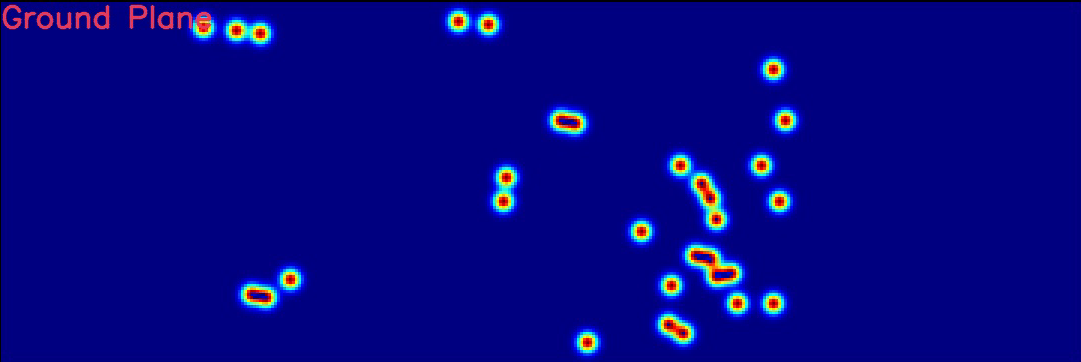}
         \caption{Ground plane heatmap (benign)}
         \label{gp_clean}
     \end{subfigure}
     \begin{subfigure}[b]{0.24\textwidth}
         \centering
         \includegraphics[width=\textwidth, height= 2.36cm]{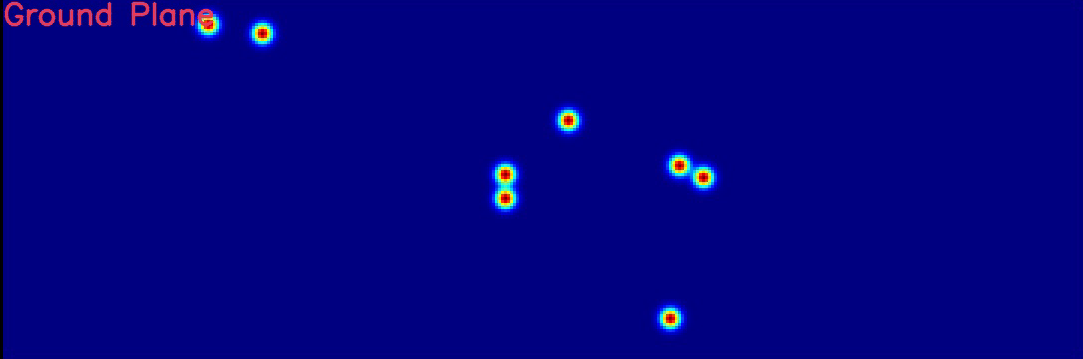}
         \caption{Ground plane heatmap (Patch)}
         \label{gp_patch}
     \end{subfigure}
        \caption{Sample of the effectiveness of our multiview patch on MVDET}
        \label{fig_ground}
\end{figure*}

%% file: sec/4_mvdetr_attack.tex
\section{Attention-aware multiview adversarial patch}
\label{sec:approach_2}

\subsection{Patch generation process}
Previous results show that the patch generated for multiview detection system fails to transfer to MVDeTr. We believe this is mainly due to the difference in architecture between the two detectors. In fact, MVDeTr builds upon MVDET by including a shadow transformer that processes the extracted features in order to account for the distortion effects introduced by the geometric transformation in the ground plane projection step. Transformers are shown to be robust against CNN-dedicated adversarial noise as shown in \cite{cnnVSviT}.

In this section, we propose an adaptive patch generation method which takes the transformer specificity into account. Since the shadow transformer introduces an attention mechanism, we propose a new loss function that adds a new attention-oriented loss function as a target to optimize during the detector's training process in addition of the regular detection loss. The attention loss is defined as follows: 
\begin{equation}
     \mathscr{L}_{Att}=\sum_{l=1}^L \sum_{h=1}^H \frac{1}{l_Q D K} \sum_{q \in Q} \sum_{d=1}^D \sum_{k=1}^K\left(P_q+P_{lhq d k} - L_k^t\right)^2
     \label{eq:att_loss}
\end{equation}
Where: 
\begin{itemize}
    \item $L$ is the number of layers,
    \item $H$ is the number of attention heads,
    \item $Q$ is the set of queries, with a size $l_Q$
    \item $D$ is the number of views in the multiview system,
    \item $K$ is the number of pointers,
    \item $L$ is the number of layers
\end{itemize}

The global loss function is then calculated by combining $\mathscr{L}_{Att}$ and $ \mathscr{L}_{multiview}$ (as defined in Equation \ref{eq:mult_loss}) using PCGrad \cite{pcgrad} to ensure that the gradients are balanced for efficient training, even when the gradients seem to point in conflicting directions.
The full patch generation algorithm is outlined in Algorithm \ref{alg2}

\begin{algorithm}
    \caption{Attention aware multiview patch generation algorithm}
    \label{alg2}
    \begin{algorithmic}[1]
        \State \textbf{Input:} {$patch\_mask$: Mask of patch position, $I_1,...,I_N$ input image for view $1,...,N$, }
        \State \textbf{Output:} {$\delta$: Attention aware adversarial multiview patch}
        \State \textit{/*Initialize the patch */}
        \State { $\delta$ $\xleftarrow{}$ \textit{{patchInit}}(gray,psize)}
        \For {$e \in [1,n_{epochs}]$}
            \For {$it \in [1,n_{iter}]$}
                \State \textit{/* Place the patch on the images using the mask*/}
                \State { $I_1,...,I_N$ $\xleftarrow{}$ \textit{{placePatch}}($I_1,...,I_N$ , $patch\_mask$, $\delta$)}
                \State \textit{/* Run the detector and get the detection loss */}
                \State {$\mathscr{L}_{multiview}$ $\xleftarrow{}$ \textit{{det($I_1,...,I_N$)}}} 
                \State \textit{/* Calculate the attention loss */}
                \State {$\mathscr{L}_{att}$ $\xleftarrow{}$ \textit{AttLoss($ptr$)}}
                \State \textit{/* Combine the losses using PCGrad */}
                \State {$\mathscr{L}_{total}$ $\xleftarrow{}$ \textit{PCGrad($\mathscr{L}_{multiview},\mathscr{L}_{Att}$)}}
                \State \textit{/* Backpropagate the loss and project the gradient to each view */}
                \State { $\nabla I_1,...,\nabla I_N$ $\xleftarrow{}$ $\frac{\partial \mathscr{L}_{total}}{\partial I_1} ,..., \frac{\partial \mathscr{L}_{total}}{\partial I_N}$ }
                \State \textit{/* Calculate the patch update */}
                \State {$step$ $\xleftarrow{}$ $\frac{\sum_{d=1}^{D} \nabla I_1 \otimes patch\_mask}{D}$}
                \State \textit{/* Update the patch*/}
                \State {$\delta (it+1)$ $\xleftarrow{}$ $\delta (it+1)$  + $step$}
            \EndFor
        \EndFor
    \end{algorithmic}
\end{algorithm}

Unlike the previous multiview patch generation method which places a patch on each target, this method places a single patch on each view. After running the detection using the adversarial inputs, we calculate the detection loss and attention loss using the detection results and the attention pointers as outlined in equations \ref{eq:mult_loss} and \ref{eq:att_loss} respectively. The two losses are then combined together using PCGrad and subsequently backpropagated and projected to each view according to equation \ref{eq:proj_grad}. To obtain the update to the patch, we filter the relevant parts of the obtained gradients using masks calculated during patch placement, and aggregate them by calculating their mean. The final step of the patch generation loop is to update the patch using the calculated update step.

The end result of this process is an adversarial patch that is able to scatter the attention vectors of the transformer to irrelevant area while simultaneously degrading the detection ability of the victim detector

\subsection{Experimental results}
\textbf{Experimental Setup:} We use a similar experimental setup to the one shown in the previous experiment: We combine the MVDeTr \cite{mvdetr} multiview object detector with the Wildtrack \cite{wildtrack} dataset. We apply our patch to each view and report the MODA and Recall.

\begin{figure}[h]
    \centering
    \includegraphics[width = \columnwidth ]{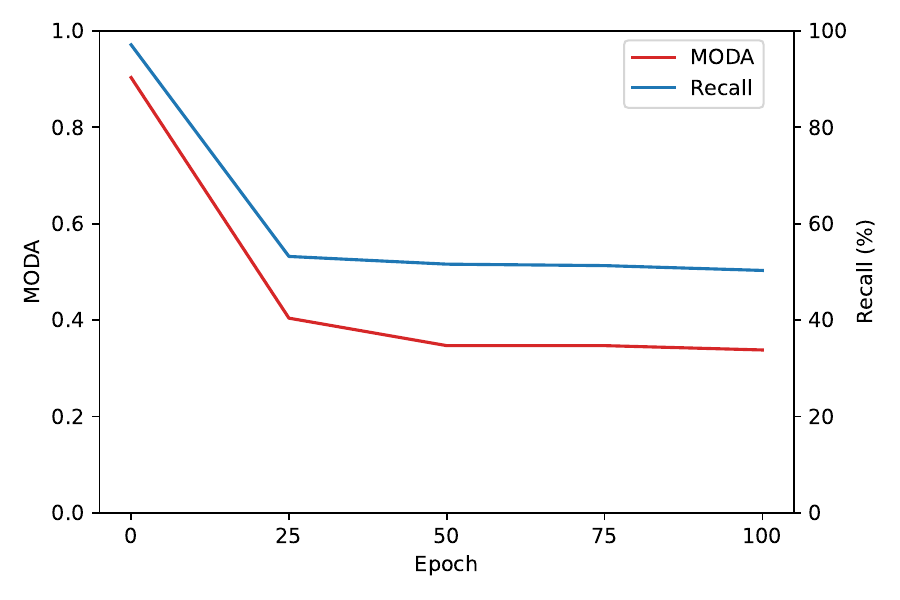}
    \caption{Results of our proposed attention aware multiview patch against MVDeTr}
    \label{fig:res_mvdetr}
\end{figure}

Figure \ref{fig:res_mvdetr} shows the evolution of the Recall and MODA training during the evolution of patch training. Our attention aware multiview patch was able to significantly degrade MVDeTr's performance with a $-62.57\%$ reduction in MODA and a $-46.09\%$ reduction in recall.

The results show that our proposed attention aware patch is effective against MVDeTr, proving that even multiview detectors that contain non-CNN elements are vulnerable to adversarial attacks.

%% file: sec/5_discussion.tex
\section{Discussion}
\label{sec:disc}

In this paper, we present two adversarial patch attacks that are able to attack and successfully degrade the performance of multiview object detector. To our knowledge, this is the first work that produces a true multiview patch. Our first patch attack aggregates gradient data from multiple views in order to train the multiview patch, while our second attention aware patch attack includes the attention loss in its optimization objective in order to effectively attack MVDeTr which includes  a transformer in its framework.

\textbf{Vulnerability of multiview object detection:} Contrary to our preliminary observations that have shown that multiview object detectors possess partial protection against existing single view adversarial patches,  our evaluation has shown that this protection is bypassed by a dedicated attacker that accounts for inter-view information sharing and feature aggregation in their attack generation framework:
\begin{itemize}
    \item Initially, the impact of existing patches on MVDET was limited: its performance was reduced by only $10\% - 30\%$ , however our patch had a significant impact on the results of MVDET with an attack success rate of $73\%$
    \item MVDeTr was impervious to the single view patches and even the MVDET patch, with a negligible performance impact of less than $1\%$. Our attention aware patch was able severely damage the performance of the detector, with an attack success rate of $62\%$
\end{itemize}

\textbf{Transferability issues:} Our MVDET attack was successful against its targeted detector but it had lost all of its effectiveness when transferred to MVDeTr. This is mainly due to the differences between the detector frameworks and the inclusion of non-CNN elements in these frameworks and raises the question of transferability issues when designing an attack against multiview object detection. These issues need to be resolved in order to reach a true universal multiview patch attack, however this is a harder task when compared to attacks against single view detectors, as multiview detectors are more complex and adopt many varieties of methods to perform the inter-view data fusion.

%% file: sec/6_related.tex
\section{Related work}
\label{sec:related}

Adversarial attacks on CNNs have been proven possible by Biggio et al. \cite{biggio}, and Szegedy et al. \cite{szegedy_noise} proposed a method to generate an adversarial noise that when added to an input image, was able to fool classifiers. Numerous attacks have been proposed using the same approach of adding a nearly imperceptible noise to the input image, such as the Fast Gradient Sign Method \cite{FGSM}, Basic iterative Method \cite{BIM}, Carlini and Wagner attacks \cite{cw}, DeepFool \cite{deepfool}, One-pixel attacks \cite{onepixel} and more. 

These threats were demonstrated to be of concern by Kurakin et al. \cite{BIM} by printing adversarial samples on paper and showing that they can still fool classifiers even through the lens of a mobile phone. Evtimov et al. \cite{evt_stop} have studied this attack concept further by managing to fool road sign detectors using printed disturbed stop sign stickers placed on to of the original sticker. Athlaye et al. \cite{eot} propose an 'Expectation Over Transformation' framework in order to expand the scope of these attacks to a noise that can be added to an arbitrary 3D object, vastly increasing the adversarial noise's effectiveness in different unfavorable conditions such as different view angles. 
 
A further development of adversarial attacks is the adversarial patch: Instead of attacking the whole image with adversarial noise, the attack is restricted to a limited area, but in exchange, the amplitude constraint is removed. The main advantage of these attacks is in the simplicity of their real life implementation: These attacks can easily be printed in a "patch" form and attached to any surface in view of the targeted detector's camera. Furthermore, these patches are universal: they are not specific to a certain input image or class, but can fool a classifier no matter the type of input image used. 

The first "Adversarial Patch" was proposed by Brown et al \cite{advpatch}, where the authors mask a part of the image and replace it with adversarial noise to fool classifier towards the targeted wrong class. The authors use a modified variant of the Expectation Over Transformation variant framework to apply rotation and scaling operations to the patch. The patch is trained using a set of images and uses gradient descent as its optimizer.

Karmon et al. \cite{lavan} introduce their LAVAN patch, in which they attempt to reduce the surface of the patch while keeping its success rates high. The authors show that even a patch that only occupies 2\% of the image area can be an effective attack. Furthermore, the authors avoid covering the main object in the input image by placing the patch in a  non-salient part of the image. The patch is optimized by maximizing the classifier's loss with regards to the correct class, thus ensuring that the output result is not correct, and by minimizing the loss with regards to the targeted class.

Thys et al. \cite{yolopatch} target the YOLO family of object detectors by creating a patch that can be used to hide people from CNN-based person detectors. This patch is created using an Adam optimizer whose objective is to minimize a loss function composed of a sum three weighted sub-goal loss functions: First, a non-printability score to ensure that the patch can be easily printed. Second, a total variation score to ensure a smooth patch. And finally, the objectness score of the person to hide in order to be undetected.

DPatch \cite{dpatch} targets object detectors in two different ways: It can target Region Proposal Networks by focusing all of the regions of interest toards the patch, therefore the RPN will not output the location of the object to detect. Also, the patch can attack image classification by maximizing the loss with regards to the correct class in the case of an untargeted attack, or by minimizing the loss towards the targeted class otherwise.

To generate patches that do not cause suspicion, especially by humans who can easily notice computer-generated patches, Hu et al. \cite{naturalistic} propose their Naturalistic Patch, that attempts to emulate the appearance of a natural image. To do so, the authors combine Generative Adversarial Networks with adversarial gradients into a network that can generate adversarial patches that are inconspicuous when found in commonly seen objects while keeping high attack performance. The authors also augment their patch with transformations such as scaling, rotations, blurring or occlusions.

Another category of adversarial attacks target 3D objects by creating an adversarial texture or mesh and applying it to a 3D object. These attacks aim to improve the results against single view detectors especially when dealing with view angle changes. However, these attacks generally do not consider information sharing across views, which is an integral component to multiview detection, therefore we can consider our proposed patch attacks to be a separate category to these attacks. Furthermore, these attacks are impractical to implement in real life: Unlike a patch with a simple geometrical shape that is easy to produce and apply anywhere, these textures are complex and have to contend with extra difficulties in manufacturing and using the patch in real life that makes them more challenging to implement beyond the virtual 3D space.

Maesumi et al. \cite{3dcloak} enhance attacks against 2D detectors by attacking them from 3D space: Using a library of varied 3D human bodies and poses, the authors generate meshes placed with different angles and distances facing the camera and use them to calculate the surface on which they train the adversarial noise and apply to the 3D model. This noise is trained by backpropagating the detection loss across the whole framework in a similar manner to regular adversarial attacks, as the framework is fully differentiable. 

Wang et al. \cite{3dcar} use the full surface of a vehicle as a platform to train and place adversarial noise that can attack a camera at any angle or distance: Using a photo-realistic 3d render of a car along with images of the car in various angles, the authors calculate the visible area on which noise can be generate by applying image segmentation on the image and rendering the noise on the model, the adversarial noise is then trained by backpropagating a loss function that combines bounding box precision, objectness score and classification accuracy.

Duan et al. \cite{3dcoating} coat a 3D object in adversarial noise in order to hide it from detection or induce misclassification: The authors generate a varied set of training images of the object using its 3D model and a collection of diverse backgrounds, and map the 3D coating from the model to the images using a number of transformations. For each image , the authors attack the all of the prominent RPN proposals simultaneously in order to minimize the cross entropy loss between the classification and the targeted class.

To fool facial recognition algorithms, Yang et al. \cite{3dface} design an adversarial textured 3D mesh that can be placed on top a person's face. A 3D reconstruction model extracts information about the input face and generates the area on which the mesh is placed. To improve performance the authors train their attack be backpropagating the loss across a low dimensional coefficient space, as it improves transferability and avoids local optimum traps. The authors propose the ability to attack RGB, depth and infrared modalities and demonstrate the effectiveness of their attack in the real-worlds by 3D printing the mask.

%% file: sec/7_conclusion.tex
\section{Conclusion}
\label{sec:conc}

In this paper, we present two adversarial patch attacks that target multiview object detectors by including information sharing across views during the patch generation process. The first patch attack aggregates gradient data from all of the views to generate the adversarial patch, and the second patch attack optimizes for attention loss in addition to the detection loss in order to attack multiview detectors that include transformers in their framework. The results of our experiments prove that multiview object detection is vulnerable to adversarial attack, despite a preliminary investigation that has shown partial robustness against existing single view adversarial patches. In light of these results, it is important to continue studying adversarial attacks in the context of multiview object detection:  On the attack side, there are still some transferability issues to solve in order to attain a universal attack. And on the defense side, increasing the robustness of multiview object detectors against adversarial attacks has become a necessity. One potential path to explore is extending existing single view defenses to work with multiview detectors.